\documentclass[conference]{IEEEtran}
\IEEEoverridecommandlockouts
\usepackage{cite}
\usepackage{amsmath,amssymb,amsfonts}
\usepackage{algorithmic}
\usepackage{graphicx}
\usepackage{textcomp}
\usepackage{xcolor}
\usepackage{tikz}
\usepackage{hyperref}
\usetikzlibrary{positioning}
\usepackage{multirow}
\usepackage{tabularx}
\usepackage{hhline}
\usepackage{xcolor}
\usepackage{subcaption}

\def\BibTeX{{\rm B\kern-.05em{\sc i\kern-.025em b}\kern-.08em
    T\kern-.1667em\lower.7ex\hbox{E}\kern-.125emX}}
\begin{document}

\title{
A Multimodal Lightweight Approach to Fault Diagnosis of Induction Motors in High-Dimensional Dataset

}

\author{\IEEEauthorblockN{Usman Ali}
\textit{GIFT University}\\
usmanali@gift.edu.pk}



\maketitle

\begin{abstract}

An accurate AI-based diagnostic system for induction motors (IMs) holds the potential to enhance proactive maintenance, mitigating unplanned downtime and curbing overall maintenance costs within an industrial environment. Notably, among the prevalent faults in IMs, a Broken Rotor Bar (BRB) fault is frequently encountered. Researchers have proposed various fault diagnosis approaches using signal processing (SP), machine learning (ML), deep learning (DL), and hybrid architectures for BRB faults. One limitation in the existing literature is the training of these architectures on relatively small datasets, risking overfitting when implementing such systems in industrial environments. This paper addresses this limitation by implementing large-scale data of BRB faults by using a transfer-learning-based lightweight DL model named ShuffleNetV2 for diagnosing one, two, three, and four BRB faults using current and vibration signal data. Spectral images for training and testing are generated using a Short-Time Fourier Transform (STFT). The dataset comprises 57,500 images, with 47,500 used for training and 10,000 for testing. Remarkably, the ShuffleNetV2 model exhibited superior performance, in less computational cost as well as accurately classifying 98.856\% of spectral images. To further enhance the visualization of harmonic sidebands resulting from broken bars, Fast Fourier Transform (FFT) is applied to current and vibration data. The paper also provides insights into the training and testing times for each model, contributing to a comprehensive understanding of the proposed fault diagnosis methodology.
 The findings of our research provide valuable insights into the performance and efficiency of different ML and DL models, offering a foundation for the development of robust fault diagnosis systems for induction motors in industrial settings.

\end{abstract}

\begin{IEEEkeywords}
induction motor, transfer learning, lightweight deep learning models, fault diagnosis, Spectral Images
\end{IEEEkeywords}

\section{Introduction}
\begin{figure}[htbp]
    \centering
    \includegraphics[width=0.46\textwidth]{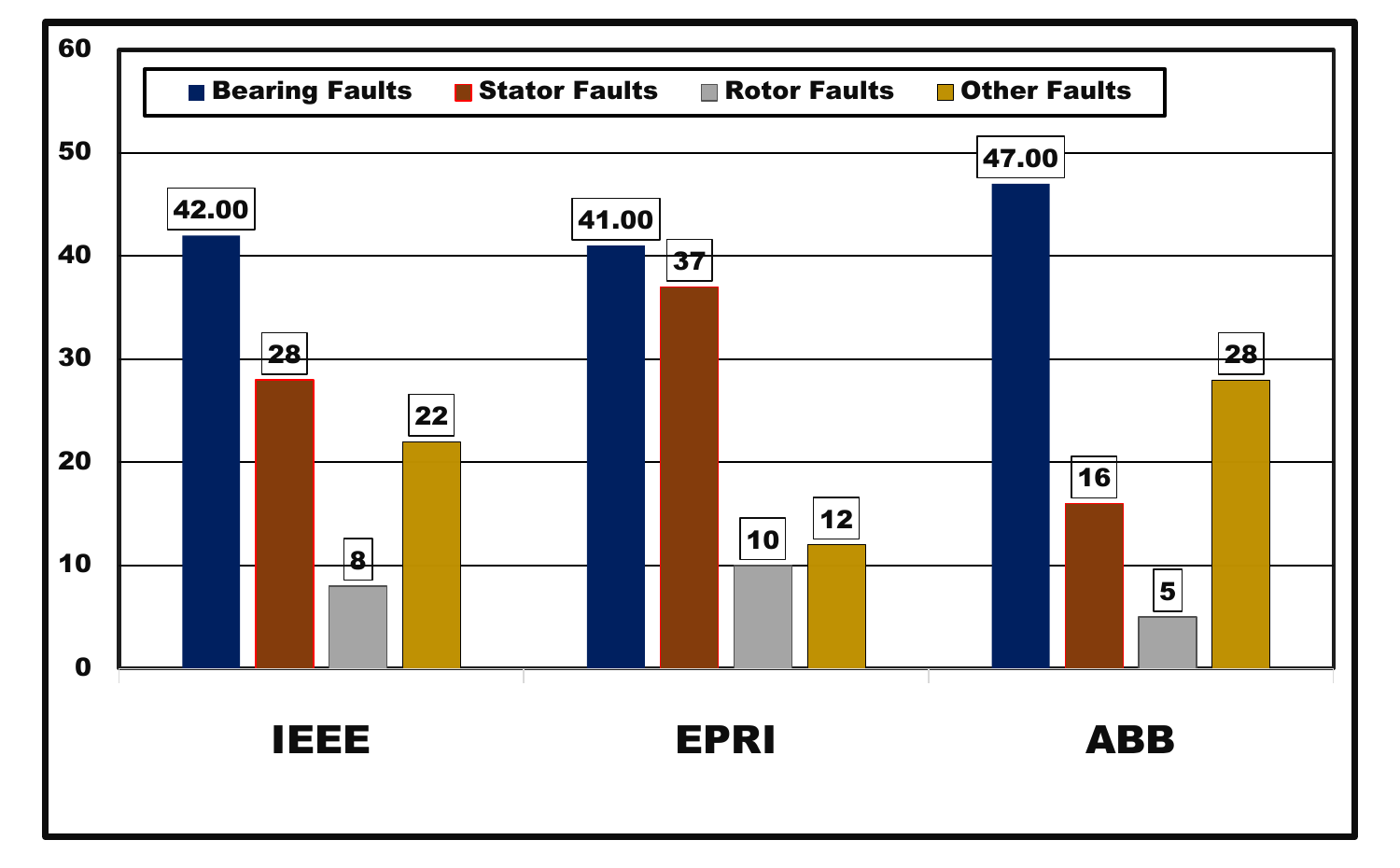} 
    \caption{Common Faults in IMs}
    \label{fig1}
\end{figure}

IMs serve as integral electromechanical components within the industrial sector, primarily employed in the fields of production, energy generation, and transport due to their inexpensive and ruggedness\cite{ali2020towards}. In recent years, extensive studies have been undertaken in the area of fault identification and classification for induction motors, underscoring their crucial role in diverse sectors\cite{b2}. Faults in IMs lead to prolonged downtimes, resulting in significant losses due to maintenance expenses and revenue reduction. These types of faults are classified as either electrical or mechanical. Electrical faults primarily occur in the rotor and stator and mechanical faults are associated with bearings and eccentricity\cite{b3}. These faults can be measured by analyzing the IMs' current, voltage, and vibration signals. Typically, the accuracy of fault classification hinges on selecting the appropriate signal and employing data collection techniques that offer vital insights into the motor's condition. Current monitoring and vibration signal measurements are predominantly employed for induction motors to achieve precision due to their non-intrusive nature and resilience\cite{b4}. According to IEEE, EPRI\cite{b5}, and ABB\cite{b6} the distributions of common faults are shown in figure \ref{fig1}. 
In this paper, we delve into the analysis of three major faults encountered in IMs: bearing, stator, and rotor faults. The bearing consists of inner and outer races supporting motor components. Issues like wear or misalignment affect operation. Stator is stationary, housing winding coils with intercoil and inter-turn spaces. Rotor, the rotating part, has isolated iron core prone to cracks due to various stresses and conditions. Defective bars generate sidebands at equal distances from the fundamental frequency. \cite{ali2023test}.   

In recent years, multiple methodologies have been employed for detecting and categorizing bearing, stator, and BRB faults in IMs. These encompass thermal, induced voltage, variation of torque, vibration, and motor current signature analysis (MCSA)\cite{b7},\cite{b8}. Similarly, various ML \cite{u1} and DL\cite{u3} methods have been used to identify and classify these faults in IMs. These included logistic regression (LR), k-nearest Neighbors (KNN), support vector machines (SVM), decision trees (DT), ensemble learning\cite{u3}, artificial neural networks (ANN)\cite{u4}, convolution neural networks (CNN), and recurrent neural networks (RNN) \cite{ali2020towards}.

Ying et al. introduced a fluid model to identify broken rotor bars by inducing thermal stress in asymmetrical squirrel cage bars of IMs\cite{b10}. Zen et al. developed an enhanced cyclostationary vibration analysis method, leveraging STFT for improved resolution and utilizing continuous wavelet transformations to extract fault patterns with high precision, this approach proved effective in diagnosing a single BRB under various load conditions\cite{b11}. In \cite{b13}, the researchers employed the hilbert transformation (HT), FFT, and ANN techniques to diagnose BRB faults in indirect variable frequency drive induction motors (IMs). The FFT is applied to extract the frequency spectrum of the stator current. Subsequently, the obtained weights were input into an ANN to enhance the model's performance and accuracy.In\cite{b14}, the authors presented research that examined contemporary BRB detection methods applicable to both line and inverter-fed IMs. The analysis emphasized crucial features and conducted comparisons among the various techniques reported in the literature. Sudip et al. implemented a finite element model to examine the rotor faults due to the transient start-up of IMs by applying the inverse thresholding in the time-frequency spectrogram\cite{b14}.In \cite{b16}, the researchers introduced the dragon transformation, a technique designed to identify the trajectory of the bar breakage frequency spectrum in the time-frequency domain. Wagner et al. presented four distinct pattern identification approaches encapsulating four ML techniques i.e., SVM, KNN, multi-layer perceptron, and a fuzzy ARTMAP, utilized for the binary and multi-classification of BRB defects in inverter-fed IMs. For the binary and multi-class classification, the implemented techniques predicted the model accuracy at 90\% and 95\% respectively on 1274 experimental samples.\cite{b17}. In\cite{b18}, the author employed multi-level feature extraction techniques, including Discrete Wavelet Transform (DWT) and binary signature combined with nearest component analysis. These extracted features then applied to the SVM and KNN classifier algorithms, resulting in a commendable success rate of 99.8\%. However, a notable limitation of this method is the risk of overfitting, given that the model trained on a relatively small dataset.In\cite{b19}, the authors employed gradient histograms to derive parameter weights from the three-phase current of the IMs. Subsequently, these measured features were applied to train a multi-layer ANN enabling the model to discern intricate patterns and relationships within the data. However, the model's performance assessment was conducted on a relatively limited dataset of 229 experimental samples, resulting in an accuracy of 95\%. Li et al. introduced a methodology based on  SVM to diagnose gearbox conditions. They successfully identified issues such as broken bars, missing teeth, and cracked gears, achieving an average accuracy at 91\%\cite{b20}Siyu et al. implemented an advanced diagnostic approach employing a pre-trained VGG-16 deep neural network to effectively identify gearbox and bearing faults in induction motors (IMs). To enhance the model's discriminative capabilities, the researchers applied wavelet transformation to the time series data, extracting crucial time-frequency features. These features were subsequently utilized to fine-tune the VGG-16 model, which underwent training with 6000 samples for gearbox faults and 5000 samples for bearing faults. Remarkably, the model demonstrated a high accuracy of 99.8\%. Nevertheless, it is important to acknowledge two inherent limitations in the implemented system. Firstly, the VGG-16 model, while powerful, is not considered lightweight. Secondly, there exists a potential overfitting concern owing to the relatively modest size of the dataset employed in the training process\cite{b21}. Shafi et al. employed a novel approach by implementing greedy-gradient graph-based semi-supervised learning to identify both binary and multi-class faults in IMs. The process involved utilizing ten data DWT windows, each comprising 9000 data samples and applying curve-fitting techniques to extract essential data features. Remarkably, their model demonstrated an impressive accuracy of 97\%. However, it is essential to note a limitation in the implemented architecture, which lies in the time complexity associated with predicting the output label class. This indicates a potential area for improvement in terms of computational efficiency to predict the output class labels for real-time applications\cite{b22}. In\cite{b24}, the researchers implemented three models—CNN, unidirectional LSTM, and bidirectional LSTM—to forecast rotor faults. The outcomes indicated that CNN exhibited superior performance compared to the other models. Notably, the paper provides a noteworthy aspect by explicitly detailing the time complexities of all the implemented models. Sajal et al. conducted a comprehensive investigation leveraging an open-source vibration dataset of broken bar faults of IMs. They employed different variants of CNNs to analyze the intricate patterns within the data. Notably, the features of the data extracted using the STFT, enrich the representation of the input for the subsequent neural network models. Upon thorough evaluation, the results demonstrated performance by the VGG-16 model Impressively, the model achieved an accuracy rate of 97\%\cite{b25}. Kevin et al. utilized six distinct CNN-based architectures, including VGG16, Inception V4, NasNETMobile, ResNet152, and SENet154, to conduct a multi-class classification of induction motors (IMs). Among these architectures, the VGG-16 model demonstrated notable performance on the small experimental dataset, comprising 16,050 samples, achieving an impressive accuracy of 99.8\%, accurately predicting the class labels\cite{b23}.

The literature discussed above indicates that researchers have utilized a variety of SP, ML, and DL techniques on limited datasets for diagnosing BRB faults in IMs. Building upon these significant advancements, this paper proposes the adoption of a lightweight CNN-based architecture, specifically known as ShuffleNet V2, for multi-class fault identification and classification. This proposed methodology is applied to a large dataset comprising stator current and vibration data from IMs. The paper outlines the following objectives and contributions.   
\begin{itemize}
    \item Frequency harmonic spectrum analysis has been executed by applying FFT to both current and vibration signals. This process allows for the visualization of surrounding sidebands attributed to broken rotor bars. The dataset utilized for this analysis is accessible on the IEEE data port\cite{b26}.
    \item  We propose the utilization of a lightweight CNN-based architecture known as ShuffleNet V2. The primary objective is to diagnose BRB faults, leveraging a substantial dataset encompassing stator current and vibration signals derived from IMs.
    \item The STFT image processing technique is employed to extract time-frequency domain features from both signals. Utilizing these image-based weights significantly enhanced the performance of the CNN-based architecture.
    \item Conducting a thorough comparative analysis, this study evaluates diverse CNN architectures within the context of a fault detection system. The assessment takes into account critical factors such as time complexity, training loss, and classification efficiency providing a comprehensive overview of their performance characteristics.
    \end{itemize}
The subsequent sections of this paper are outlined as follows: Section II presents the theoretical background on BRB faults and CNN-based architectures. Section III delves into the data acquisition process, while Section IV outlines the proposed methodology. Section V is dedicated to the discussion and presentation of experimental results. Finally, Section VI concludes our work.

\section{Theoretical Background}

\subsection{Fast Fourier Transform (FFT)}
The FFT is a computational algorithm specifically crafted for computing the Discrete Fourier Transform (DFT) of a given signal. Fourier analysis involves the conversion of a signal from its original domain, usually time, into a representation within the frequency domain. The DFT is obtained by decomposing a sequence of values into constituent components characterized by different frequencies. If we have a sequence of complex numbers represented as $x_0, x_1, x_2, \ldots, x_{n-1}$, each $x_i$ denotes an element in the sequence, where $i$ ranges from 0 to $n-1$. The formula defining DFT is expressed in equation \ref{eq1}: 

\begin{equation}
X(j) = \sum_{n=0}^{N-1} x(n) \cdot e^{-\frac{2\pi i}{N}jn} \label{eq1}   
\end{equation}

\begin{figure}[http]
    \centering
    \includegraphics[width=0.45\textwidth]{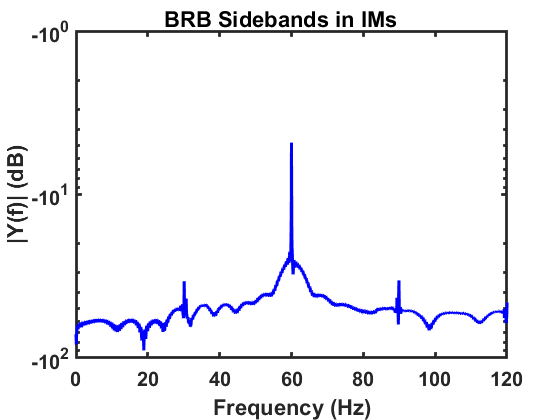} 
    \caption{FFT Spectrum of BRB in IMs}
    \label{fig2}
\end{figure}

In this context, \(X(j)\) denotes the \(j\)-th frequency component within the frequency domain. \(x(n)\) represents the input sequence in the time domain, where \(N\) stands for the number of points in the sequence. The symbol \(i\) corresponds to the imaginary unit.

Now, the frequency component of  rotor bar \(f_{\text{brb}}\) can be calculated using equation  \ref{eq2}:

\begin{equation}
f_{\text{brb}} = f_{\text{s}} \cdot (1 \pm 2ks) \label{eq2}  
\end{equation}

The right and left sideband harmonic spectrum of the implemented BRB fault dataset is shown in figure \ref{fig2}.

\subsection{Short-Time Fourier Transform (STFT)}
The STFT is a method for analyzing signals that allow the observation of changes in the frequency content of a signal over time.
This method proves particularly valuable when dealing with non-stationary signals characterized by varying frequency components with time\cite{b27}.The discrete STFT of a discrete signal \(x[z]\) with a window function \(w[n]\) is shown in equation (3):
\begin{equation}
X[m, i] = \sum_{z=0}^{z-1} x[z] \cdot w[z - mR] \cdot e^{-j 2\pi i z / Z} 
\end{equation}
\begin{figure}[http]
    \centering
    \includegraphics[width=0.45\textwidth]{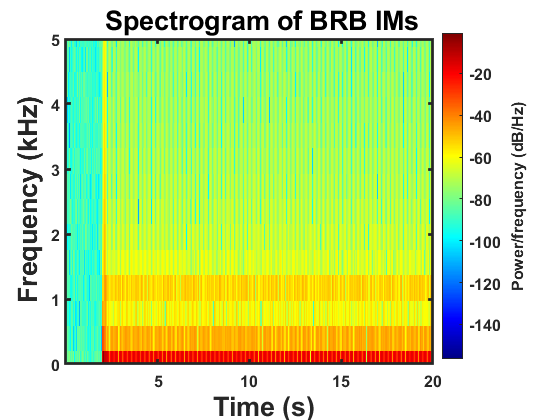} 
    \caption{STFT Spectrum of BRB in IMs}
    \label{fig3}
\end{figure}

\begin{figure*}[http]
    \centering
    \includegraphics[width=0.95\textwidth]{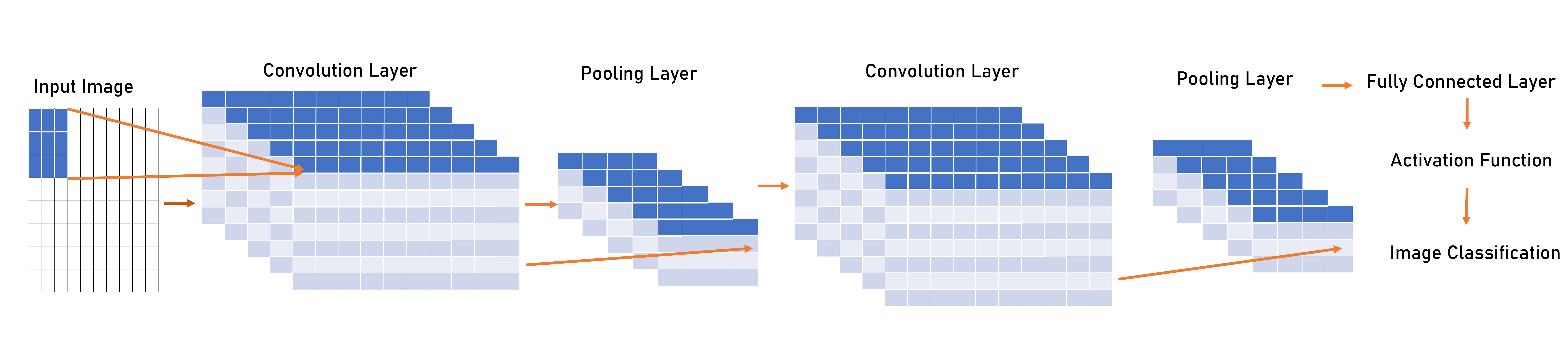} 
    \caption{CNN Architecture}
    \label{fig4}
\end{figure*}

Where, \(X[m, i]\) is the STFT coefficient at index \(m\) and frequency index \(i\), \(x[z]\) is the input signal, \(w[z - mR]\) is the window function centered at time index \(m\) with a hop size of \(R\) between successive windows, \(Z\) is the length of the DFT  for each window.

The discrete STFT is computed by analyzing short segments of the signal using the window function and applying the discrete fourier transform to each segment is shown in figure \ref{fig3}.

\subsection{Convolutional Neural Network (CNN)}

The STFT-based images are utilized to construct a CNN for the automated diagnosis of IMs. The basic CNN architecture is shown in figure\ref{fig4}.
The convolution kernel filter is applied to the input image, computing the element-wise multiplication, and summing at each position. Mathematically, the 2D convolution operation is defined in equation \ref{eq4}:
\begin{equation}
 (I * K)(i, j) = \sum_m \sum_n I(i+m, j+n) \cdot K(m, n) \label{eq4}
\end{equation}
Here, \(I(i, j)\) represents the pixel intensity at position \((i, j)\) in the input image, and \(K(m, n)\) represents the filter coefficient at position \((m, n)\).
The Rectified Linear Unit (ReLU) serves as a widely employed activation function in CNN. It imparts non-linearity to the network by transforming all negative input values to zero. Mathematically, the ReLU function is defined in equation \ref{eq5}:
\begin{equation}
\text{ReLU}(w) = \max(0, w) \label{eq5}
\end{equation}

Pooling layers reduce spatial dimensions by selecting the maximum value from a group of neighboring pixels. Mathematically, max pooling is defined in equation \ref{eq6}:
\begin{equation}
\text{MaxPooling}(w) = \max(w_{i, j}, w_{i, j+1}, w_{i+1, j}, w_{i+1, j+1})   \label{eq6} 
\end{equation}

Fully connected layers perform high-level reasoning. The output \(y_i\) for each neuron is computed using weights \(z_{ij}\), biases \(b_i\), and input values \(w_j\):
\begin{equation}
y_i = \text{ReLU}\left(\sum_j z_{ij} \cdot w_j + b_i\right) \label{eq7}
\end{equation}

The training of CNNs involves minimizing a loss function by adjusting the weights and biases through back-propagation. This process uses gradient descent or a variant to update the parameters.

\subsection{Evaluation Measures}
We employed accuracy (A{c}), precision (P{r}), recall (R{c}), and F-1 scores as evaluation metrics to assess the model's performance.
A{c} is the proportion of accurately predicted instances relative to the total samples in the dataset can derived using equation \ref{eq8}.
\begin{equation}
 A{c} = \frac{\text{TP+TN}}{\text{TP+TN+FP+FN}} \label{eq8}
\end{equation}
P{r} assesses the accuracy of positive predictions by determining the ratio of correctly predicted positive instances to the total instances predicted as positive. This is calculated using the equation \ref{eq9}.
\begin{equation}
    P{r} = \frac{\text{TP}}{\text{TP} + \text{FP}} \label{eq9}
\end{equation}
 R{c} represents the proportion of accurately predicted positive instances among all labels belonging to the actual positive class to be calculated using equation \ref{eq10}.
\begin{equation}
    R{c} = \frac{\text{TP}}{\text{TP} + \text{FN}} \label{eq10}
\end{equation}
The f-1 score, which balances precision and recall through the harmonic mean, is computed using equation \ref{eq11}.
\begin{equation}
    \text{F-1 score} = \frac{2 * (\text{Pr} \times \text{Rc})}{\text{Pr} + \text{Rc}} \label{eq11}
\end{equation}

\section{Dataset Description}
In our study, we utilized an open-source dataset accessible on the IEEE data port, published by Treml et al. titled "Experimental Database for Detecting and Diagnosing Rotor Broken Bars in 3-Phase IMs"\cite{b26}.
\begin{figure}[htbp]
    \centering
    \includegraphics[width=0.45\textwidth]{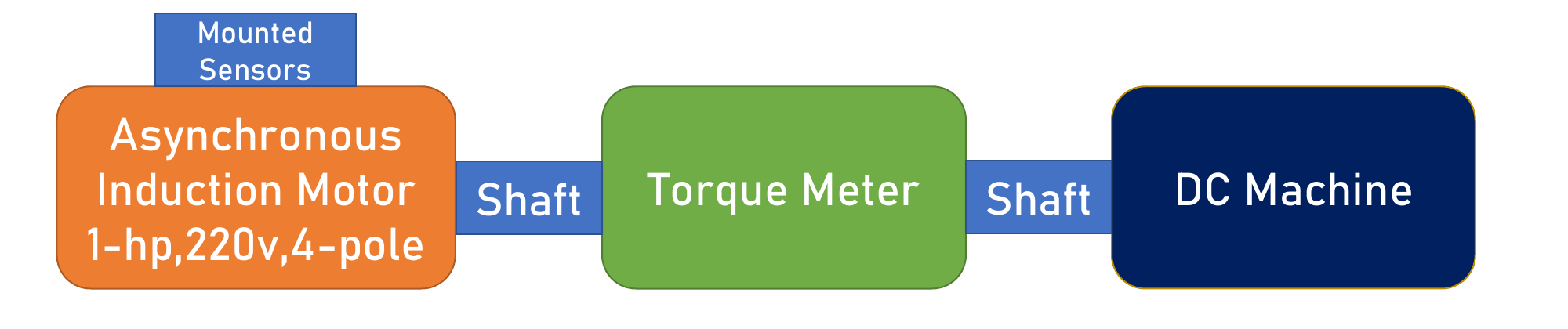} 
    \caption{Block Diagram of Induction Motor Test Bench }
    \label{fig5}
\end{figure}
\begin{figure*}[http]
    \centering
    \includegraphics[width=0.95\textwidth]{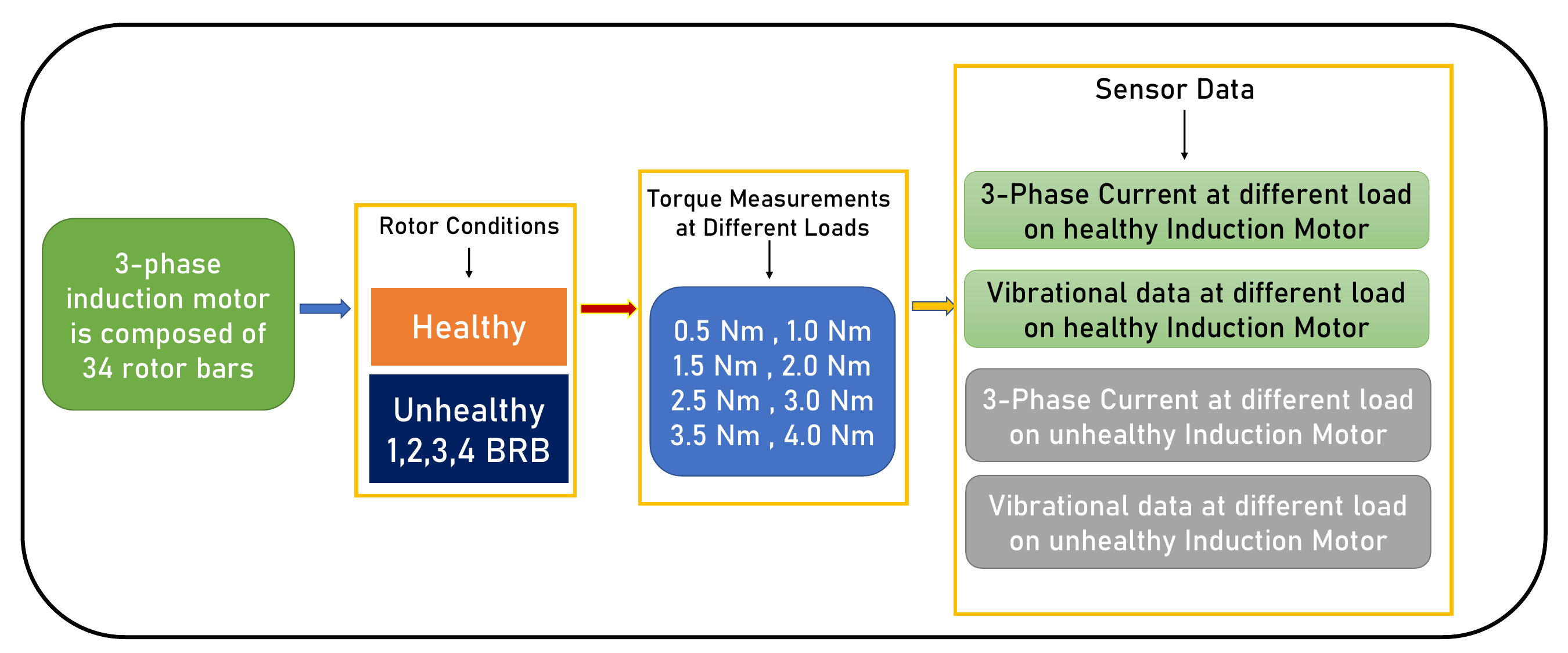} 
    \caption{Dataset Description Flow}
    \label{fig6}
\end{figure*}
\subsection{Experimental Setup}
The experimental setup is constructed with a 3-phase IM intricately synchronized to a DC machine, acting as a generator to emulate load torque as shown in figure \ref{fig5}.

This connection is facilitated by a shaft that utilizes a rotary torque for precise control.
The dataset includes electromechanical signals produced from experiments conducted on 3-phase IMs. These experiments involved variation of mechanical loads on the IM axis and introducing various levels of BRB defects in the motor rotor, encompassing data related to rotors without defects.

\subsubsection{Testing Motor Properties}
The IM is rated at 1 horsepower, operates at voltages of 220V/380V, draws currents of 3.02A/1.75A, features 4 poles, runs at a frequency of 60 Hz, and operates at a speed of 1715 rpm. The rotor, constructed in the squirrel cage type, consists of 34 bars. The regulation of the load torque involves adjusting the field winding voltage of a DC generator, a process accomplished by utilizing a single-phase voltage variation that is equipped with a filtered full-bridge rectifier. The IM underwent testing across a range encompassing 12.5\%, 25\%, 37.5\%, 50\%, 62.5\%, 75\%, 87.5\%, and 100\% of its full load capacity.

\subsubsection{Mounted Sensor details}
For the electrical signal measurements, the currents were precisely assessed utilizing alternating current probes, specifically designed meters with a capability of up to 50ARMS. These probes, corresponding to precision meters, feature an output voltage of 10 mV/A. Additionally, voltages are directly calculated at the induction terminals using voltage points on the oscilloscope and this measurement equipment was sourced from the manufacturer Yokogawa.
For the assessment of mechanical signals, a set of five axial accelerometers were concurrently employed. These accelerometers exhibit an acuteness of 10 mV/mm/s, possess a frequency range spanning between 5 to 2000Hz, and feature stainless steel housing. This configuration enables comprehensive vibration measurements on both the drive end and non-drive end sides of the motor, accommodating axial or radial orientations in both horizontal and vertical directions.

Simultaneous sampling of all signals occurred over a consistent duration of 18 seconds for each loading condition. The experimental procedure encompassed ten repetitions, capturing the transient to steady-state phases of the induction motor. The recorded details of the 3-phase BRBs dataset are shown in figure\ref{fig6}.

\begin{figure*}
    \centering
    \includegraphics[width=0.99\textwidth]{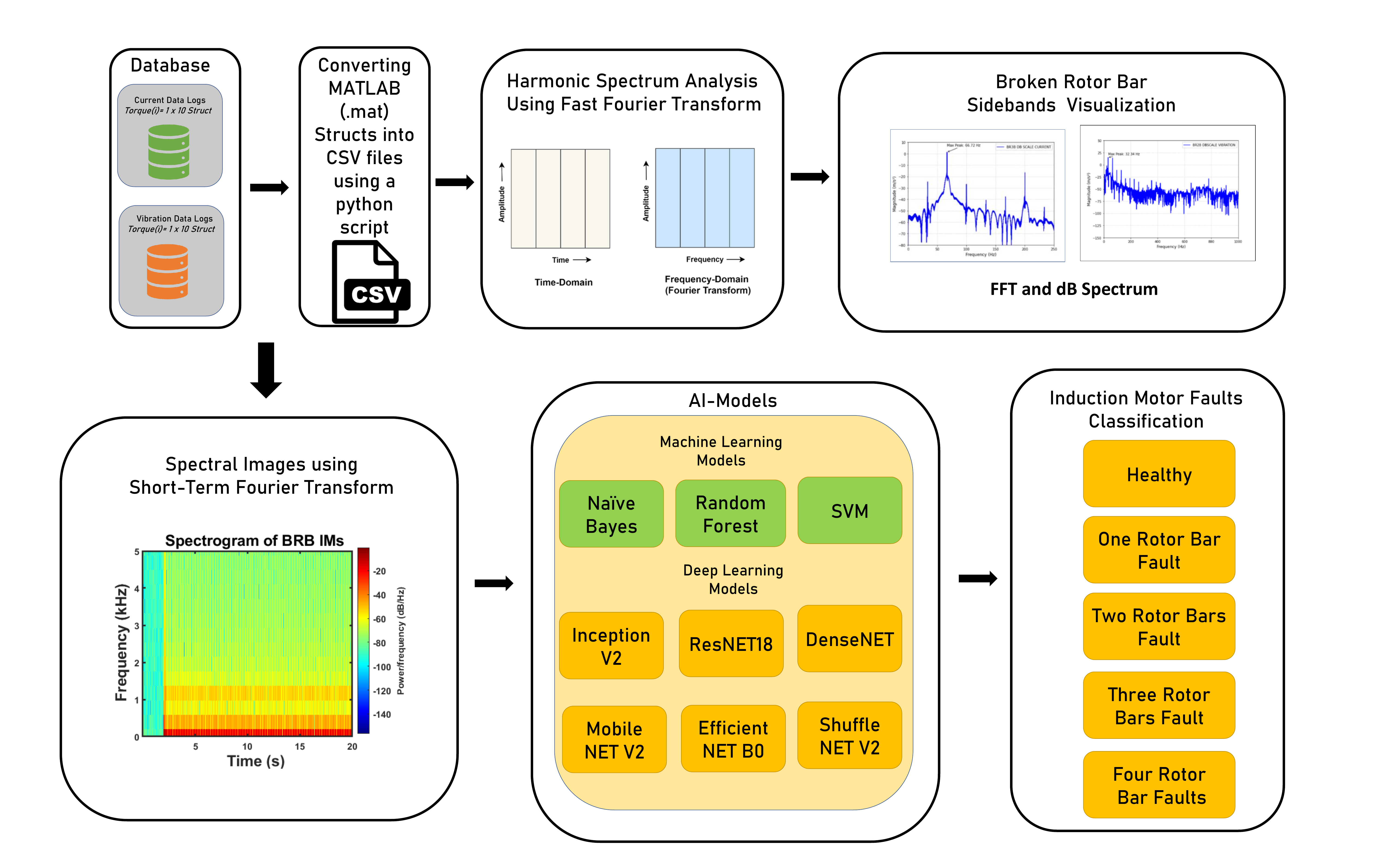} 
    \caption{Step-by-Step Working Methodology}
    \label{fig7}
\end{figure*}

\section{Proposed Methodology}
Figure \ref{fig7} illustrates the proposed working diagram, depicting the utilization of data presented in the form of stator current and vibration signals, conveniently provided in .mat format. These datasets are organized in a structural array format within MATLAB. The structural format is subsequently transformed into CSV files through the utilization of a Python script, enhancing compatibility for further processing and analysis. An effective SP technique proves instrumental in extracting valuable features from a signal\cite{b28}. We applied both FFT and STFT techniques for feature extraction. The FFT technique is used to visualize the harmonic spectrum of both current and vibration signals. Simultaneously, the STFT technique is applied to generate spectral images, subsequently aiding the neural network in training its parametric weights. A comprehensive exploration of ML methodologies involved the application of a range of algorithms, such as Naïve
Bayes, Random
Forest, and SVM  to spectral image datasets originating from both current and vibration signals. Despite this diversity, these conventional techniques demonstrated limited efficacy when confronted with the intricate image data specific to IMs. Recognizing the need for heightened accuracy in image dataset analysis, CNN emerged as the preeminent choice, leveraging its specialized architecture to capture intricate patterns and relationships within the spectral images for more precise classification outcomes\cite{b29}. Within the framework of CNN, the initial step involves the application of convolution using an array of kernels or filters. This is followed by the incorporation of a non-linear activation function, along with the integration of batch normalization to enhance the stability and efficiency of the learning process. Additionally, pooling may be implemented if deemed necessary, contributing to the extraction of relevant features and reduction of spatial dimensions in the network's hierarchical architecture. 
In the context of this study, fault diagnosis is undertaken using CNN-based pre-trained transfer learning models. Transfer learning, a methodology involving the utilization of a model initially trained on a specific task and subsequently repurposed for a related task, played a pivotal role in this approach. The CNN operations are executed by implementing Transfer Learning, harnessing various pre-trained models such as  InceptionV2, DenseNET, MobileNEtV2, ResNEt18, efficientNET, and ShuffleNEtV2. 
This array of models serves the purpose of feature extraction from the spectrogram images. The extracted features are subsequently transmitted to fully connected networks, culminating in a comprehensive integration of spectral information. Ultimately,a softmax function is applied to facilitate the multi-class classification task, ensuring the assignment of probabilities to different fault categories based on the learned features.

The proposed steps for diagnosing a BRB fault in an IM are summarized as follows:
\begin{itemize}
    \item The extensive dataset, encompassing a substantial volume of stator current and vibrational signals recorded under diverse torque conditions, has been stored in a structural array format within a .mat file. This information is now being transformed into CSV files for further analysis and accessibility.
    \item The SP method, specifically FFT is implemented on both time domain datasets. This process aims to visually represent the sideband harmonics surrounding the fundamental frequency, providing domain experts with a comprehensive view of the harmonic spectrum. This visualization aids experts in making precise predictions by enhancing their understanding of the harmonic characteristics.  
    \item The data undergoes an image processing technique, specifically STFT, to produce time-frequency domain spectral images. These images are subsequently employed in the training of neural networks, contributing to the adjustment of their weights for enhanced performance.
    \begin{figure*}[http] 
    \centering
    \includegraphics[width=0.95\textwidth, height=0.75\textwidth]{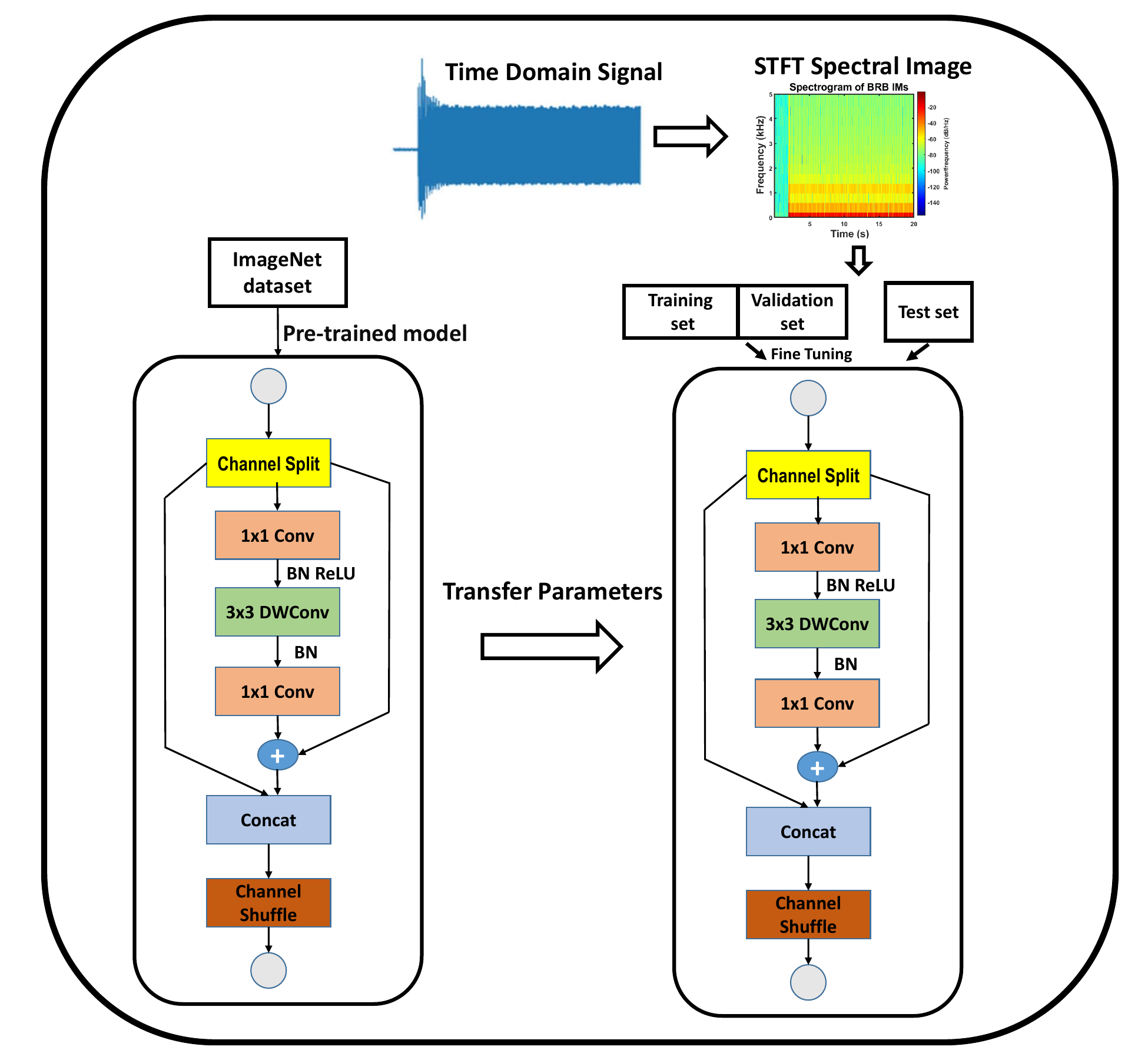} 
    \caption{Fine Tune Transfer Learning ShuffleNETV2 Approach }
    \label{fig8}
    \end{figure*}
    \item Each image dataset derived from the measurements of current and vibration signals is systematically categorized and labeled as "healthy motor (HLT)", "one broken bar (BRB1)", "two broken bars (BRB2)", "three broken bars (BRB3)", "four broken bars (BRB4)".
    \item The categorized spectrograms are utilized in the application of several ML algorithms, including  Naive Bayes,
    Random Forest, and SVM. In addition, diverse CNN architectures, namely InceptionV2, DenseNET, MobileNEtV2, ResNEt18, efficientNET, and ShuffleNEtV2 have been implemented to address the image classification task.  
    \begin{table}
\centering
\caption{Spectral Images corresponding to each label}
\label{tab1}
\begin{tabular}{|c|c|c|}
\hline
\textbf{Rotor Condition}   & \textbf{Class Label} & \textbf{\# of Spectrograms} \\ \hline
HLT              & 0                    & 11500                           \\ \hline
BRB1  & 1                   & 11500                           \\ \hline
BRB2 & 2                    & 11500                           \\ \hline
BRB3 & 3                    & 11500                           \\ \hline
BRB4 & 4                    & 11500                           \\ \hline
\end{tabular}
\end{table}
    \item The analysis reveals that the ShuffleNet V2, a lightweight transfer learning CNN-based model, surpasses all other models in performance. This superiority is attributed to its optimized architecture, reduced training time complexity, and remarkable accuracy in predicting correct labels across a substantial dataset of IM images. The proposed lightweight architecture of ShuffleNet-V2 
    is visually presented in figure \ref{fig8}.  
    \item As presented in Table \ref{tab1}, a comprehensive array of spectral images aligns with each distinct label. In this context, an impressive total of 57500 images has been generated for each output label. This expansive dataset plays a pivotal role in augmenting the precision of output predictions and fostering resilience against overfitting, thereby mitigating the incidence of false negatives. 
\end{itemize}   
 \section{Results and Discussion}

In this section, we present the experimental results conducted on both healthy and unhealthy squirrel cage IMs. A comprehensive description of the hardware and software environment employed in our research is imperative for contextualizing the experimental outcomes are shown in Table \ref{tab2} . 

\begin{table}[htbp]
    \centering
    \caption{System Configuration}
    \label{tab2}
    \begin{tabular}{|clclcl|}
        \hline
        \multicolumn{1}{|c|}{\multirow{4}{*}{\textbf{Hardware Specifications}}} & \multicolumn{1}{l|}{\textbf{CPU}} & \multicolumn{2}{l|}{Intel(R) @3.20 GHz(32 CPU)} \\ \cline{2-4} 
        \multicolumn{1}{|c|}{} & \multicolumn{1}{l|}{\textbf{GPU}} & \multicolumn{2}{l|}{NVIDIA RTX 3080 Ti 12GB} \\ \cline{2-4} 
        \multicolumn{1}{|c|}{} & \multicolumn{1}{l|}{\textbf{RAM}} & \multicolumn{2}{l|}{64 GB DDR4} \\ \cline{2-4} 
        \multicolumn{1}{|c|}{} & \multicolumn{1}{l|}{\textbf{Storage}} & \multicolumn{2}{l|}{256 SSD + 1TB HDD} \\ \hline
        \multicolumn{1}{|l|}{\multirow{3}{*}{\textbf{Software Environment}}} & \multicolumn{1}{l|}{\textbf{OS}} & \multicolumn{2}{l|}{Windows 11 Pro 64-bits} \\ \cline{2-4} 
        \multicolumn{1}{|l|}{} & \multicolumn{1}{l|}{\textbf{Languages}} & \multicolumn{2}{l|}{Python 3.9, MATLAB R2021a} \\ \cline{2-4} 
        \multicolumn{1}{|l|}{} & \multicolumn{1}{l|}{\textbf{Libraries}} & \multicolumn{2}{l|}{PyTorch 2.2.0 + CUDA 11.8} \\ \hline
    \end{tabular}
\end{table}

The first part of the results presented the visualization of the sideband distribution using the harmonics spectrum of various BRB faults. For that purpose, we employed FFT on the current and vibration signals to observe the surrounding sidebands 

\begin{figure*}[http] 
    \centering
    \includegraphics[width=0.95\textwidth]{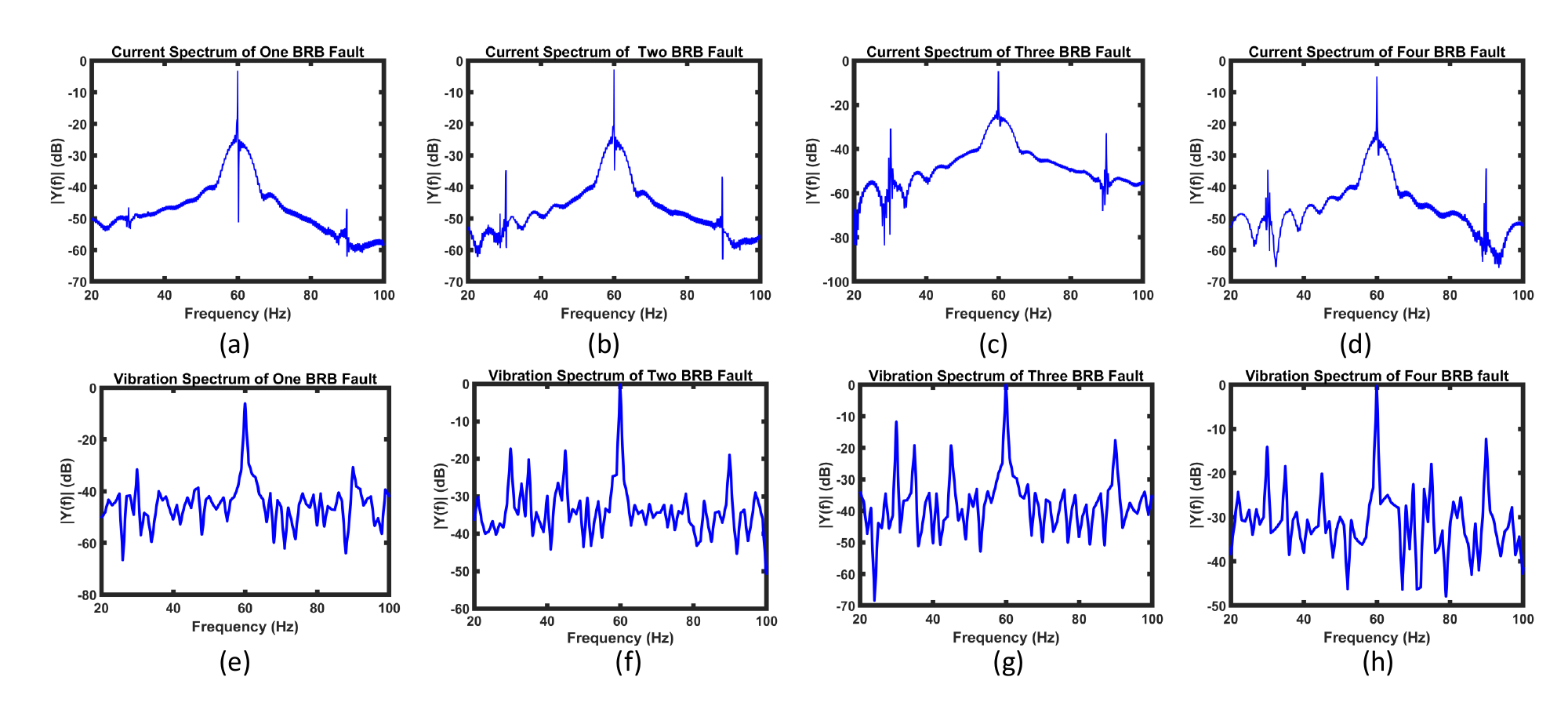} 
    \caption{FFT Harmonic Spectrum Analysis of Current and Vibration Data Captured from IMs with BRB Fault}
    \label{fig22}    
\end{figure*}

\begin{table}[]
\centering
\caption{ML Models Test Accuracy}
\label{tab3}
\begin{tabular}{|l|c|}
\hline
\multicolumn{1}{|c|}{\textbf{Model Name}} & \multicolumn{1}{l|}{\textbf{Model Test Accuracy (\%)}} \\ \hline
\textbf{Naive Bayes}                      & 42                                                     \\ \hline
\textbf{Random Forest}                    & 64                                                     \\ \hline
\textbf{SVM}                              & 57                                                     \\ \hline
\end{tabular}
\end{table}

\begin{table}[]
\centering
\captionsetup{justification=centering} 
  \caption{Multi-class Classification Report on Different DL Models}
\label{tab4}
\begin{tabular}{|c|c|ccc|}
\hline
\multirow{2}{*}{\textbf{Model Name}}     & \multirow{2}{*}{\textbf{\begin{tabular}[c]{@{}c@{}}Motor \\ Condition\end{tabular}}} & \multicolumn{3}{c|}{\textbf{Classification Report}}                                                \\ \cline{3-5} 
                                         &                                                                                      & \multicolumn{1}{c|}{\textbf{Precision}} & \multicolumn{1}{c|}{\textbf{Recall}} & \textbf{F1-Score} \\ \hline
\multirow{6}{*}{\textbf{Inception V2}}   & \textbf{HLT}                                                                         & \multicolumn{1}{c|}{1.00}               & \multicolumn{1}{c|}{0.93}            & 0.95              \\ \cline{2-5} 
                                         & \textbf{BRB1}                                                                        & \multicolumn{1}{c|}{0.98}               & \multicolumn{1}{c|}{0.93}            & 0.95              \\ \cline{2-5} 
                                         & \textbf{BRB2}                                                                        & \multicolumn{1}{c|}{0.96}               & \multicolumn{1}{c|}{0.93}            & 0.95              \\ \cline{2-5} 
                                         & \textbf{BRB3}                                                                        & \multicolumn{1}{c|}{0.92}               & \multicolumn{1}{c|}{0.97}            & 0.94              \\ \cline{2-5} 
                                         & \textbf{BRB4}                                                                        & \multicolumn{1}{c|}{0.94}               & \multicolumn{1}{c|}{0.95}            & 0.94              \\ \cline{2-5} 
                                         & \textbf{Accuracy}                                                                    & \multicolumn{1}{c|}{}                   & \multicolumn{1}{c|}{}                & 0.9562            \\ \hline
\multirow{6}{*}{\textbf{ResNET18}}       & \textbf{HLT}                                                                         & \multicolumn{1}{c|}{1.00}               & \multicolumn{1}{c|}{1.00}            & 1.00              \\ \cline{2-5} 
                                         & \textbf{BRB1}                                                                        & \multicolumn{1}{c|}{0.96}               & \multicolumn{1}{c|}{0.91}            & 0.94              \\ \cline{2-5} 
                                         & \textbf{BRB2}                                                                        & \multicolumn{1}{c|}{0.97}               & \multicolumn{1}{c|}{0.93}            & 0.95              \\ \cline{2-5} 
                                         & \textbf{BRB3}                                                                        & \multicolumn{1}{c|}{0.92}               & \multicolumn{1}{c|}{0.97}            & 0.95              \\ \cline{2-5} 
                                         & \textbf{BRB4}                                                                        & \multicolumn{1}{c|}{0.91}               & \multicolumn{1}{c|}{0.95}            & 0.93              \\ \cline{2-5} 
                                         & \textbf{Accuracy}                                                                    & \multicolumn{1}{c|}{}                   & \multicolumn{1}{c|}{}                & 0.9514            \\ \hline
\multirow{6}{*}{\textbf{DenseNET}}       & \textbf{HLT}                                                                         & \multicolumn{1}{c|}{1.00}               & \multicolumn{1}{c|}{1.00}            & 1.00              \\ \cline{2-5} 
                                         & \textbf{BRB1}                                                                        & \multicolumn{1}{c|}{0.95}               & \multicolumn{1}{c|}{0.97}            & 0.91              \\ \cline{2-5} 
                                         & \textbf{BRB2}                                                                        & \multicolumn{1}{c|}{0.97}               & \multicolumn{1}{c|}{0.91}            & 0.94              \\ \cline{2-5} 
                                         & \textbf{BRB3}                                                                        & \multicolumn{1}{c|}{0.93}               & \multicolumn{1}{c|}{0.97}            & 0.95              \\ \cline{2-5} 
                                         & \textbf{BRB4}                                                                        & \multicolumn{1}{c|}{0.86}               & \multicolumn{1}{c|}{0.96}            & 0.91              \\ \cline{2-5} 
                                         & \textbf{Accuracy}                                                                    & \multicolumn{1}{c|}{}                   & \multicolumn{1}{c|}{}                & 0.9419            \\ \hline
\multirow{6}{*}{\textbf{MobileNetV2}}   & \textbf{BRB1}                                                                        & \multicolumn{1}{c|}{0.94}               & \multicolumn{1}{c|}{0.93}            & 0.94              \\ \cline{2-5} 
                                         & \textbf{BRB2}                                                                        & \multicolumn{1}{c|}{0.98}               & \multicolumn{1}{c|}{0.91}            & 0.94              \\ \cline{2-5} 
                                         & \textbf{BRB3}                                                                        & \multicolumn{1}{c|}{0.93}               & \multicolumn{1}{c|}{0.96}            & 0.95              \\ \cline{2-5} 
                                         & \textbf{BRB4}                                                                        & \multicolumn{1}{c|}{0.92}               & \multicolumn{1}{c|}{0.96}            & 0.94              \\ \cline{2-5} 
                                         & \textbf{HLT}                                                                     & \multicolumn{1}{c|}{1.00}               & \multicolumn{1}{c|}{1.00}            & 1.00              \\ \cline{2-5} 
                                         & \textbf{Accuracy}                                                                    & \multicolumn{1}{c|}{}                   & \multicolumn{1}{c|}{}                & 0.9530            \\ \hline
\multirow{6}{*}{\textbf{EfficientNetB0}} & \textbf{BRB1}                                                                        & \multicolumn{1}{c|}{0.92}               & \multicolumn{1}{c|}{0.95}            & 0.94              \\ \cline{2-5} 
                                         & \textbf{BRB2}                                                                        & \multicolumn{1}{c|}{0.97}               & \multicolumn{1}{c|}{0.92}            & 0.94              \\ \cline{2-5} 
                                         & \textbf{BRB3}                                                                        & \multicolumn{1}{c|}{0.93}               & \multicolumn{1}{c|}{0.96}            & 0.95              \\ \cline{2-5} 
                                         & \textbf{BRB4}                                                                        & \multicolumn{1}{c|}{0.92}               & \multicolumn{1}{c|}{0.96}            & 0.94              \\ \cline{2-5} 
                                         & \textbf{HLT}                                                                     & \multicolumn{1}{c|}{1.00}               & \multicolumn{1}{c|}{1.00}            & 1.00              \\ \cline{2-5} 
                                         & \textbf{Accuracy}                                                                    & \multicolumn{1}{c|}{}                   & \multicolumn{1}{c|}{}                & 0.9535            \\ \hline
\multirow{6}{*}{\textbf{ShuffleNetV2}}   & \textbf{BRB1}                                                                        & \multicolumn{1}{c|}{0.99}               & \multicolumn{1}{c|}{0.95}            & 0.97              \\ \cline{2-5} 
                                         & \textbf{BRB2}                                                                        & \multicolumn{1}{c|}{0.99}               & \multicolumn{1}{c|}{0.99}            & 0.99              \\ \cline{2-5} 
                                         & \textbf{BRB3}                                                                        & \multicolumn{1}{c|}{0.99}               & \multicolumn{1}{c|}{1.00}            & 0.99              \\ \cline{2-5} 
                                         & \textbf{BRB4}                                                                        & \multicolumn{1}{c|}{0.96}               & \multicolumn{1}{c|}{0.99}            & 0.97              \\ \cline{2-5} 
                                         & \textbf{HLT}                                                                     & \multicolumn{1}{c|}{1.00}               & \multicolumn{1}{c|}{1.00}            & 1.00              \\ \cline{2-5} 
                                         & \textbf{Accuracy}                                                                    & \multicolumn{1}{c|}{}                   & \multicolumn{1}{c|}{}                & 0.9885           \\ \hline
\end{tabular}
\end{table}

\begin{table*}
\centering
\caption{Multi-Class Confusion Matrix}
\label{tab5}
\begin{tabular}{|c|cccccc|c|cccccc|}
\hline
\multirow{3}{*}{\textbf{(1)}}                                                      & \multicolumn{6}{c|}{\textbf{Inception V2}}                                                                                                                                                              & \multirow{3}{*}{\textbf{(2)}}                                                      & \multicolumn{6}{c|}{\textbf{ResNET18}}                                                                                                                                                                 \\ \cline{2-7} \cline{9-14} 
                                                                                   & \multicolumn{6}{c|}{\textbf{Predicted Labels}}                                                                                                                                                          &                                                                                    & \multicolumn{6}{c|}{\textbf{Predicted Labels}}                                                                                                                                                         \\ \cline{2-7} \cline{9-14} 
                                                                                   & \multicolumn{1}{c|}{}              & \multicolumn{1}{c|}{\textbf{BRB1}}    & \multicolumn{1}{c|}{\textbf{BRB2}} & \multicolumn{1}{c|}{\textbf{BRB3}} & \multicolumn{1}{c|}{\textbf{BRB4}} & \textbf{HLT}  &                                                                                    & \multicolumn{1}{c|}{}              & \multicolumn{1}{c|}{\textbf{BRB1}}    & \multicolumn{1}{c|}{\textbf{BRB2}} & \multicolumn{1}{c|}{\textbf{BRB3}} & \multicolumn{1}{c|}{\textbf{BRB4}} & \textbf{HLT} \\ \hline
\multirow{5}{*}{\textbf{\begin{tabular}[c]{@{}c@{}}Actual\\  Labels\end{tabular}}} & \multicolumn{1}{c|}{\textbf{BRB1}}    & \multicolumn{1}{c|}{\textbf{\textcolor{blue}{1850}}} & \multicolumn{1}{c|}{20}            & \multicolumn{1}{c|}{13}            & \multicolumn{1}{c|}{117}           & 0              & \multirow{5}{*}{\textbf{\begin{tabular}[c]{@{}c@{}}Actual\\  Labels\end{tabular}}} & \multicolumn{1}{c|}{\textbf{BRB1}}    & \multicolumn{1}{c|}{\textbf{\textcolor{blue}{1822}}} & \multicolumn{1}{c|}{10}            & \multicolumn{1}{c|}{21}            & \multicolumn{1}{c|}{147}           & 0             \\ \cline{2-7} \cline{9-14} 
                                                                                   & \multicolumn{1}{c|}{\textbf{BRB2}} & \multicolumn{1}{c|}{23}            & \multicolumn{1}{c|}{\textbf{\textcolor{blue}{1867}}} & \multicolumn{1}{c|}{104}           & \multicolumn{1}{c|}{6}             & 0              &                                                                                    & \multicolumn{1}{c|}{\textbf{BRB2}} & \multicolumn{1}{c|}{29}            & \multicolumn{1}{c|}{\textbf{\textcolor{blue}{1855}}} & \multicolumn{1}{c|}{97}            & \multicolumn{1}{c|}{19}            & 0             \\ \cline{2-7} \cline{9-14} 
                                                                                   & \multicolumn{1}{c|}{\textbf{BRB3}} & \multicolumn{1}{c|}{11}            & \multicolumn{1}{c|}{39}            & \multicolumn{1}{c|}{\textbf{\textcolor{blue}{1943}}} & \multicolumn{1}{c|}{7}             & 0              &                                                                                    & \multicolumn{1}{c|}{\textbf{BRB3}} & \multicolumn{1}{c|}{17}            & \multicolumn{1}{c|}{21}            & \multicolumn{1}{c|}{\textbf{\textcolor{blue}{1947}}} & \multicolumn{1}{c|}{15}            & 0             \\ \cline{2-7} \cline{9-14} 
                                                                                   & \multicolumn{1}{c|}{\textbf{BRB4}} & \multicolumn{1}{c|}{11}            & \multicolumn{1}{c|}{24}            & \multicolumn{1}{c|}{63}            & \multicolumn{1}{c|}{\textbf{\textcolor{blue}{1902}}} & 0              &                                                                                    & \multicolumn{1}{c|}{\textbf{BRB4}} & \multicolumn{1}{c|}{26}            & \multicolumn{1}{c|}{29}            & \multicolumn{1}{c|}{50}            & \multicolumn{1}{c|}{\textbf{\textcolor{blue}{1895}}} & 0             \\ \cline{2-7} \cline{9-14} 
                                                                                   & \multicolumn{1}{c|}{\textbf{HLT}} & \multicolumn{1}{c|}{0}             & \multicolumn{1}{c|}{0}             & \multicolumn{1}{c|}{0}             & \multicolumn{1}{c|}{0}             & \textbf{\textcolor{blue}{2000}}  &                                                                                    & \multicolumn{1}{c|}{\textbf{HLT}} & \multicolumn{1}{c|}{0}             & \multicolumn{1}{c|}{0}             & \multicolumn{1}{c|}{0}             & \multicolumn{1}{c|}{0}             & \textbf{\textcolor{blue}{2000}} \\ \hline
\multirow{3}{*}{\textbf{(3)}}                                                      & \multicolumn{6}{c|}{\textbf{DenseNET}}                                                                                                                                                                  & \multirow{3}{*}{\textbf{(4)}}                                                      & \multicolumn{6}{c|}{\textbf{MobileNET V2}}                                                                                                                                                             \\ \cline{2-7} \cline{9-14} 
                                                                                   & \multicolumn{6}{c|}{\textbf{Predicted Labels}}                                                                                                                                                          &                                                                                    & \multicolumn{6}{c|}{\textbf{Predicted Labels}}                                                                                                                                                         \\ \cline{2-7} \cline{9-14} 
                                                                                   & \multicolumn{1}{c|}{}              & \multicolumn{1}{c|}{\textbf{BRB1}}    & \multicolumn{1}{c|}{\textbf{BRB2}} & \multicolumn{1}{c|}{\textbf{BRB3}} & \multicolumn{1}{c|}{\textbf{BRB4}} & \textbf{HLT} &                                                                                    & \multicolumn{1}{c|}{}              & \multicolumn{1}{c|}{\textbf{BRB1}}    & \multicolumn{1}{c|}{\textbf{BRB2}} & \multicolumn{1}{c|}{\textbf{BRB3}} & \multicolumn{1}{c|}{\textbf{BRB4}} & \textbf{HLT} \\ \hline
\multirow{5}{*}{\textbf{\begin{tabular}[c]{@{}c@{}}Actual\\ Labels\end{tabular}}}  & \multicolumn{1}{c|}{\textbf{BRB1}}    & \multicolumn{1}{c|}{\textbf{\textcolor{blue}{1735}}} & \multicolumn{1}{c|}{14}            & \multicolumn{1}{c|}{4}             & \multicolumn{1}{c|}{247}           & 0              & \multirow{5}{*}{\textbf{\begin{tabular}[c]{@{}c@{}}Actual\\ Labels\end{tabular}}}  & \multicolumn{1}{c|}{\textbf{BRB1}}    & \multicolumn{1}{c|}{\textbf{\textcolor{blue}{1865}}} & \multicolumn{1}{c|}{9}             & \multicolumn{1}{c|}{12}            & \multicolumn{1}{c|}{114}           & 0             \\ \cline{2-7} \cline{9-14} 
                                                                                   & \multicolumn{1}{c|}{\textbf{BRB1}} & \multicolumn{1}{c|}{58}            & \multicolumn{1}{c|}{\textbf{\textcolor{blue}{1828}}} & \multicolumn{1}{c|}{83}            & \multicolumn{1}{c|}{31}            & 0              &                                                                                    & \multicolumn{1}{c|}{\textbf{BRB2}} & \multicolumn{1}{c|}{44}            & \multicolumn{1}{c|}{\textbf{\textcolor{blue}{1848}}} & \multicolumn{1}{c|}{78}            & \multicolumn{1}{c|}{30}            & 0             \\ \cline{2-7} \cline{9-14} 
                                                                                   & \multicolumn{1}{c|}{\textbf{BRB3}} & \multicolumn{1}{c|}{20}            & \multicolumn{1}{c|}{22}            & \multicolumn{1}{c|}{\textbf{\textcolor{blue}{1934}}} & \multicolumn{1}{c|}{24}            & 0              &                                                                                    & \multicolumn{1}{c|}{\textbf{BRB3}} & \multicolumn{1}{c|}{39}            & \multicolumn{1}{c|}{28}            & \multicolumn{1}{c|}{\textbf{\textcolor{blue}{1913}}} & \multicolumn{1}{c|}{20}            & 0             \\ \cline{2-7} \cline{9-14} 
                                                                                   & \multicolumn{1}{c|}{\textbf{BRB3}} & \multicolumn{1}{c|}{16}            & \multicolumn{1}{c|}{14}            & \multicolumn{1}{c|}{48}            & \multicolumn{1}{c|}{\textbf{\textcolor{blue}{1922}}} & 0              &                                                                                    & \multicolumn{1}{c|}{\textbf{BRB4}} & \multicolumn{1}{c|}{33}            & \multicolumn{1}{c|}{17}            & \multicolumn{1}{c|}{40}            & \multicolumn{1}{c|}{\textbf{\textcolor{blue}{1910}}} & 0             \\ \cline{2-7} \cline{9-14} 
                                                                                   & \multicolumn{1}{c|}{\textbf{HLT}} & \multicolumn{1}{c|}{0}             & \multicolumn{1}{c|}{0}             & \multicolumn{1}{c|}{0}             & \multicolumn{1}{c|}{0}             & \textbf{\textcolor{blue}{2000}}  &                                                                                    & \multicolumn{1}{c|}{\textbf{HLT}} & \multicolumn{1}{c|}{0}             & \multicolumn{1}{c|}{0}             & \multicolumn{1}{c|}{0}             & \multicolumn{1}{c|}{0}             & \textbf{\textcolor{blue}{2000}} \\ \hline
\multirow{3}{*}{\textbf{(5)}}                                                      & \multicolumn{6}{c|}{\textbf{EfficientNET}}                                                                                                                                                              & \multirow{3}{*}{\textbf{(6)}}                                                      & \multicolumn{6}{c|}{\textbf{ShuffleNET}}                                                                                                                                                               \\ \cline{2-7} \cline{9-14} 
                                                                                   & \multicolumn{6}{c|}{\textbf{Predicted Labels}}                                                                                                                                                          &                                                                                    & \multicolumn{6}{c|}{\textbf{Predicted Labels}}                                                                                                                                                         \\ \cline{2-7} \cline{9-14} 
                                                                                   & \multicolumn{1}{c|}{}              & \multicolumn{1}{c|}{\textbf{BRB1}}    & \multicolumn{1}{c|}{\textbf{BRB2}} & \multicolumn{1}{c|}{\textbf{BRB3}} & \multicolumn{1}{c|}{\textbf{BRB4}} & \textbf{HLT}  &                                                                                    & \multicolumn{1}{c|}{\textbf{}}     & \multicolumn{1}{c|}{\textbf{BRB1}}    & \multicolumn{1}{c|}{\textbf{BRB2}} & \multicolumn{1}{c|}{\textbf{BRB3}} & \multicolumn{1}{c|}{\textbf{BRB4}} & \textbf{HLT} \\ \hline
\multirow{5}{*}{\textbf{\begin{tabular}[c]{@{}c@{}}Actual \\ Labels\end{tabular}}} & \multicolumn{1}{c|}{\textbf{BRB1}}    & \multicolumn{1}{c|}{\textbf{\textcolor{blue}{1869}}} & \multicolumn{1}{c|}{8}             & \multicolumn{1}{c|}{17}            & \multicolumn{1}{c|}{106}           & 0              & \multirow{5}{*}{\textbf{\begin{tabular}[c]{@{}c@{}}Actual\\ Labels\end{tabular}}}  & \multicolumn{1}{c|}{\textbf{BRB1}}    & \multicolumn{1}{c|}{\textbf{\textcolor{blue}{1909}}} & \multicolumn{1}{c|}{2}             & \multicolumn{1}{c|}{6}             & \multicolumn{1}{c|}{83}            & 0             \\ \cline{2-7} \cline{9-14} 
                                                                                   & \multicolumn{1}{c|}{\textbf{BRB2}} & \multicolumn{1}{c|}{57}            & \multicolumn{1}{c|}{\textbf{\textcolor{blue}{1822}}} & \multicolumn{1}{c|}{82}            & \multicolumn{1}{c|}{39}            & 0              &                                                                                    & \multicolumn{1}{c|}{\textbf{BRB2}} & \multicolumn{1}{c|}{7}             & \multicolumn{1}{c|}{\textbf{\textcolor{blue}{1984}}} & \multicolumn{1}{c|}{9}             & \multicolumn{1}{c|}{0}             & 0             \\ \cline{2-7} \cline{9-14} 
                                                                                   & \multicolumn{1}{c|}{\textbf{BRB3}} & \multicolumn{1}{c|}{29}            & \multicolumn{1}{c|}{18}            & \multicolumn{1}{c|}{\textbf{\textcolor{blue}{1923}}} & \multicolumn{1}{c|}{30}            & 0              &                                                                                    & \multicolumn{1}{c|}{\textbf{BRB3}} & \multicolumn{1}{c|}{0}             & \multicolumn{1}{c|}{0}             & \multicolumn{1}{c|}{\textbf{\textcolor{blue}{1994}}} & \multicolumn{1}{c|}{6}             & 0             \\ \cline{2-7} \cline{9-14} 
                                                                                   & \multicolumn{1}{c|}{\textbf{BRB4}} & \multicolumn{1}{c|}{29}            & \multicolumn{1}{c|}{14}            & \multicolumn{1}{c|}{41}            & \multicolumn{1}{c|}{\textbf{\textcolor{blue}{1916}}} & 0              &                                                                                    & \multicolumn{1}{c|}{\textbf{BRB4}} & \multicolumn{1}{c|}{7}             & \multicolumn{1}{c|}{8}             & \multicolumn{1}{c|}{11}            & \multicolumn{1}{c|}{\textbf{\textcolor{blue}{1974}}} & 0             \\ \cline{2-7} \cline{9-14} 
                                                                                   & \multicolumn{1}{c|}{\textbf{HLT}} & \multicolumn{1}{c|}{0}             & \multicolumn{1}{c|}{0}             & \multicolumn{1}{c|}{0}             & \multicolumn{1}{c|}{0}             & \textbf{\textcolor{blue}{2000}}  &                                                                                    & \multicolumn{1}{c|}{\textbf{HLT}} & \multicolumn{1}{c|}{0}             & \multicolumn{1}{c|}{0}             & \multicolumn{1}{c|}{0}             & \multicolumn{1}{c|}{0}             & \textbf{\textcolor{blue}{2000}} \\ \hline
\end{tabular}
\end{table*}

\begin{table*}
\centering
\caption{Performance and Computational Comparative Analysis of Various Transfer Learning Models}
\label{tab6}
\begin{tabular}{|l|c|c|c|c|c|c|}
\hline
\textbf{Model Name} & \textbf{Train Time (s)} & \textbf{Memory/Epoch (MB)} & \textbf{Train Acc (\%)} & \textbf{Val Acc (\%)} & \textbf{Test Time (s)} & \textbf{Test Acc (\%)} \\ \hline
\textbf{DenseNET} & 73355 & 82.28 & 99.79 & 94.59 & 37.5 & 94.19 \\ \hline
\textbf{ResNET18} & 77228 & 128.35 & 99.87 & 94.04 & 52.13 & 95.13 \\ \hline
\textbf{Inception V2} & 50404 & 90.960 & 99.67 & 94.82 & 43.17 & 95.08 \\ \hline
\textbf{MobileNetV2} & 19164 & 26.528 & 99.28 & 94.02 & 58.40 & 95.30 \\ \hline
\textbf{EfficientNetB0} & 33670 & 47.530 & 99.70 & 94.53 & 44.485 & 95.35 \\ \hline
\textbf{ShuffleNETV2} & \textbf{\textcolor{blue}{15348}} & \textbf{\textcolor{blue}{15.085}} & \textbf{\textcolor{blue}{99.63}} & \textbf{\textcolor{blue}{98.96}} & \textbf{\textcolor{blue}{31.51}} & \textbf{\textcolor{blue}{98.85}} \\ \hline
\end{tabular}
\end{table*}

throughout the harmonic spectrum of the fundamental frequency, as depicted in Figure 9. Figures 9(a-d) illustrate the current harmonic spectrum of the unhealthy IMs, while figures 9(e-h) display the vibration spectrum of the unhealthy IMs.

Figures 9a and 9e present the single BRB spectrum of the IMs, revealing right and left sidebands at 90 Hz and 30 Hz, respectively, positioned equidistantly around the fundamental frequency of 60 Hz with lower magnitudes. Figures 9b and 9f exhibit the dual BRB faults spectrum in both current and vibration data. Notably, the vibration spectrum displays two additional sidebands around approximately 45 Hz and 25 Hz. Moving on to Figures 9c and 9g, they showcase the triple BRB faults spectrum for both current and vibration data. In this case, the sidebands maintain equal spacing but exhibit higher magnitudes. Notably, in the current-based harmonic spectrum, lower-magnitude harmonic spikes emerge around the sidebands. Finally, Figures 9d and 9h illustrate the quadruple BRB faults spectrum for both current and vibration data. Similar to the triple BRB case, the sidebands maintain equal spacing with higher magnitudes. Additionally, the vibration spectrum shows a single harmonic spike with increased amplitude at around 74 Hz.

The presentation of the harmonic spectrum for these BRB faults serves the purpose of improving the discernibility of fault patterns based on different types of BBR damage in both current and vibration data. This detailed analysis provides valuable insights for domain experts, enabling them to better understand and identify distinct fault characteristics. Furthermore, these unique features contribute to enhancing the efficiency of the model, facilitating more accurate and effective fault detection and diagnosis.


In the subsequent phase of this study, we evaluated the performance of various ML models. Initially, we trained fundamental ML-based models, namely Naive Bayes, Random Forest, and SVM, on our dataset. Naive Bayes operates by classifying images according to probabilistic principles, while Random Forest constructs an ensemble of decision trees during training, with each tree trained on a random subset of the data and features. SVM classifies images based on the principle of identifying the hyperplane that best segregates different classes in the feature space. However, these traditional models demonstrated suboptimal performance when applied to our multi-class spectral-based image dataset. The accuracies attained are outlined in Table \ref{tab3}. Naive Bayes correctly classified 42\% of images, Random Forest achieved a correct classification rate of 64\%, and SVM correctly classified 57\% of the images.

Following the evaluation of traditional ML models, we proceeded to implement six CNN-based DL models for multi-class image classification on our dataset. These models comprise well-known architectures such as Inception V2, ResNet18, and DenseNet, alongside three lightweight models: MobileNetV2, EfficientNetB0, and ShuffleNetV2.

Table \ref{tab4} presents the multi-class classification report for the evaluated models, utilizing precision, recall, F1-score, and accuracy metrics. Furthermore, Table \ref{tab5} illustrates the multi-class confusion matrix, offering insights into the performance of each model. To assess the capabilities of these models, a total of 2000 images per class—representing categories such as one BRB, two BRB, three BRB, four BRB faults , and healthy —were employed. The outcomes for each model are detailed as follows:
\begin{itemize}
    \item InceptionV2 achieved commendable results, correctly classifying 1850, 1867, 1943, 1902, and all 2000 images for the respective classes.
    \item ResNET18 demonstrated proficiency by correctly classifying 1822, 1855, 1947, 1895, and all 2000 images for the specified classes.
    \item DenseNET exhibited reliable performance, correctly classifying 1735, 1828, 1934, 1922, and all 2000 images for the respective classes.
    \item MobileNETV2 showcased effectiveness with accurate classifications of 1865, 1848, 1913, 1910, and 1999 images for the corresponding classes.
    \item EfficientNET demonstrated robust performance by correctly classifying 1869, 1822, 1923, 1916, and all 2000 images for the specified classes.
    \item ShuffleNETV2 displayed proficiency, correctly classifying 1850, 1867, 1943, 1902, and all 2000 images for the respective classes.
\end{itemize}

Table \ref{tab6} provides a comprehensive overview of the final performance evaluation of these models, alongside their computational costs. This table includes key metrics such as model training time, training accuracy, validation accuracy, test time, memory storage per epoch, and model test accuracy.

\begin{figure}[htbp]
    \centering
    \includegraphics[width=0.48\textwidth]{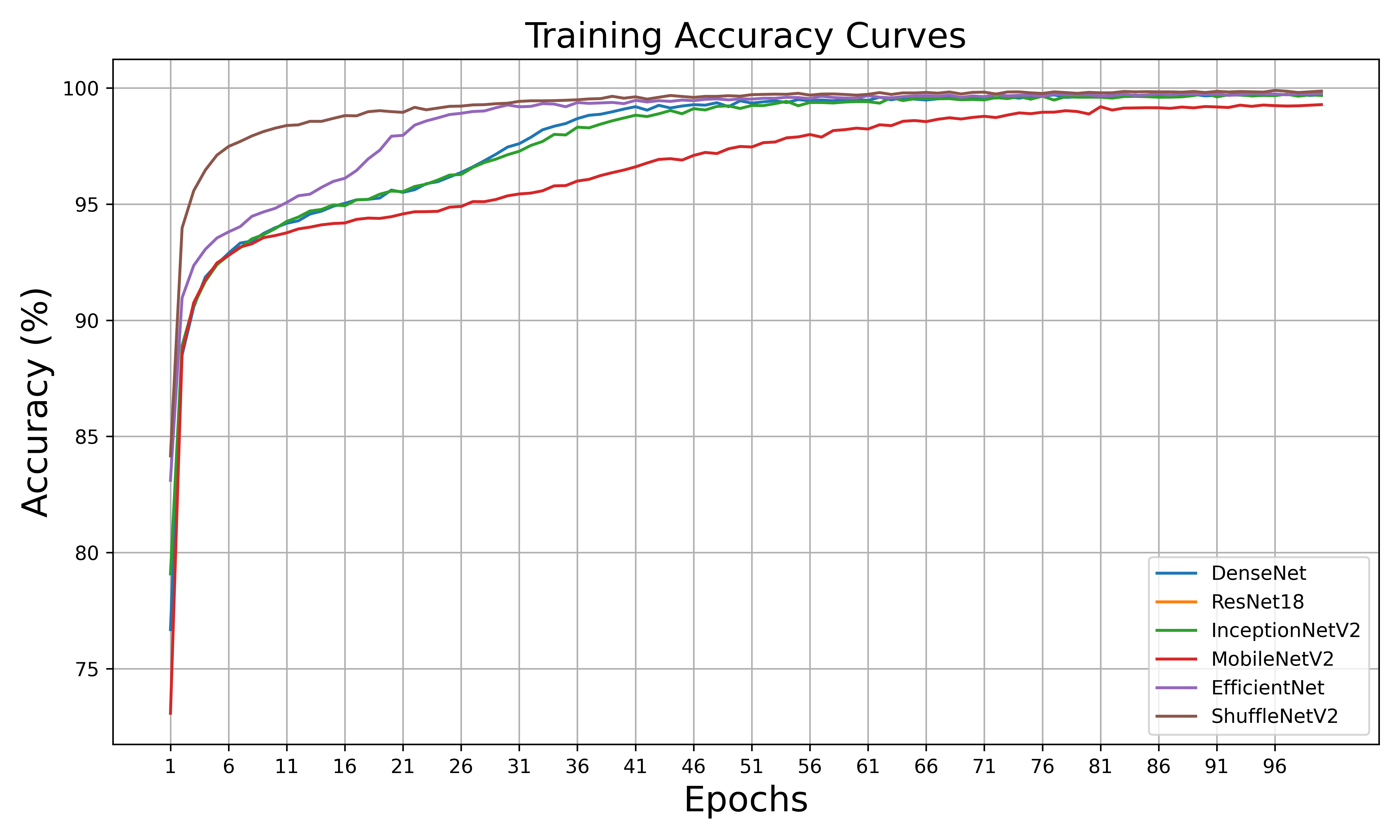} 
    \caption{Training Accuracy Curves}
    \label{fig10}
\end{figure}

\begin{figure}[htbp]
    \centering
    \includegraphics[width=0.48\textwidth]{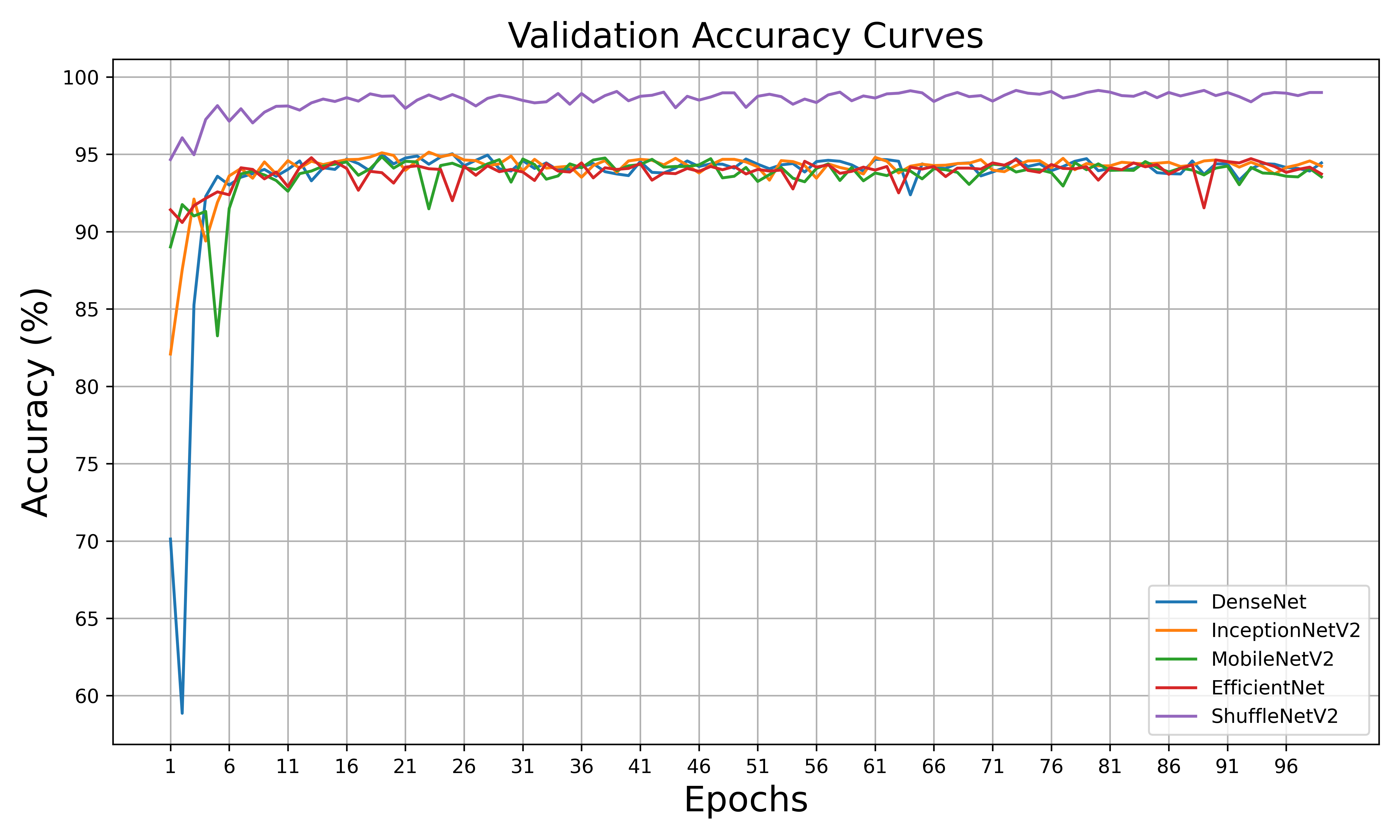} 
    \caption{Validation Accuracy Curves}
    \label{fig11}
\end{figure}

A noteworthy observation is the reduced training time of lightweight DL models compared to the initial three models. Particularly, ShuffleNETV2 stands out for its minimal training time of 15348 seconds, demonstrating efficiency in model training. Additionally, it excels in test time, requiring only 31.51 seconds, surpassing all other models in computational efficiency. Furthermore, ShuffleNETV2 showcases superior performance, accurately classifying 98.85\% of test images, thereby outperforming all other models in terms of classification accuracy. The figure illustrates the training and validation accuracies of all models in \ref{fig10} and \ref{fig11}, respectively..

\section{Conclusion  and Future Work}


In industrial settings, early fault diagnosis of IMs holds critical significance. Among the significant faults encountered in IMs, the BRB fault stands out. We applied six distinct CNN-based architectures to a comprehensive BRB fault dataset, comprising current and vibration signals. Our findings revealed that a lightweight DL architecture, namely ShuffleNETV2, surpasses all other models in terms of accuracy and computational efficiency, including training, testing, and storage costs.

Looking ahead, we envision deploying this lightweight architecture for fault diagnosis across a broader spectrum of IM issues, including bearing, stator, and air-gap eccentricity, leveraging large-scale datasets.

\bibliographystyle{IEEEtran}
\bibliography{Bibliography}

\begin{thebibliography}{10}
\providecommand{\url}[1]{#1}
\csname url@samestyle\endcsname
\providecommand{\newblock}{\relax}
\providecommand{\bibinfo}[2]{#2}
\providecommand{\BIBentrySTDinterwordspacing}{\spaceskip=0pt\relax}
\providecommand{\BIBentryALTinterwordstretchfactor}{4}
\providecommand{\BIBentryALTinterwordspacing}{\spaceskip=\fontdimen2\font plus
\BIBentryALTinterwordstretchfactor\fontdimen3\font minus \fontdimen4\font\relax}
\providecommand{\BIBforeignlanguage}[2]{{%
\expandafter\ifx\csname l@#1\endcsname\relax
\typeout{** WARNING: IEEEtran.bst: No hyphenation pattern has been}%
\typeout{** loaded for the language `#1'. Using the pattern for}%
\typeout{** the default language instead.}%
\else
\language=\csname l@#1\endcsname
\fi
#2}}
\providecommand{\BIBdecl}{\relax}
\BIBdecl

\bibitem{ali2020towards}
U.~Ali, R.~Hafiz, T.~Tauqeer, U.~Younis, W.~Ali, and A.~Ahmad, ``Towards machine learning based real-time fault identification and classification in high power induction motors,'' in \emph{2020 5th International Conference on Robotics and Automation Engineering (ICRAE)}.\hskip 1em plus 0.5em minus 0.4em\relax IEEE, 2020, pp. 46--53.

\bibitem{b2}
A.~Almounajjed, A.~K. Sahoo, and M.~K. Kumar, ``Diagnosis of stator fault severity in induction motor based on discrete wavelet analysis,'' \emph{Measurement}, vol. 182, p. 109780, 2021.

\bibitem{b3}
O.~E. Hassan, M.~Amer, A.~K. Abdelsalam, and B.~W. Williams, ``Induction motor broken rotor bar fault detection techniques based on fault signature analysis--a review,'' \emph{IET Electric Power Applications}, vol.~12, no.~7, pp. 895--907, 2018.

\bibitem{b4}
A.~Almounajjed, A.~K. Sahoo, and M.~K. Kumar, ``Diagnosis of stator fault severity in induction motor based on discrete wavelet analysis,'' \emph{Measurement}, vol. 182, p. 109780, 2021.

\bibitem{b5}
M.~E.~H. Benbouzid, ``A review of induction motors signature analysis as a medium for faults detection,'' \emph{IEEE transactions on industrial electronics}, vol.~47, no.~5, pp. 984--993, 2000.

\bibitem{b6}
P.~Mighdoll, R.~Bloss, and F.~Hayashi, ``Improved motors for utility applications. volume 2. industry assessment study. final report,'' Booz, Allen and Hamilton, Inc., Cleveland, OH (USA), Tech. Rep., 1982.

\bibitem{ali2023test}
U.~Ali, W.~Ali, M.~U. Noor, M.~U. Ramzan, M.~U. Aslam, and H.~Farooq, ``Test rig for the fault diagnosis of 3-phase small scale induction motor,'' in \emph{2023 International Conference on IT and Industrial Technologies (ICIT)}.\hskip 1em plus 0.5em minus 0.4em\relax IEEE, 2023, pp. 1--5.

\bibitem{b7}
S.~Mezani, N.~Takorabet, and B.~Laporte, ``A combined electromagnetic and thermal analysis of induction motors,'' \emph{IEEE transactions on Magnetics}, vol.~41, no.~5, pp. 1572--1575, 2005.

\bibitem{b8}
A.~Roque, J.~Calado, and J.~Ruiz, ``Vibration analysis versus current signature analysis,'' \emph{IFAC Proceedings Volumes}, vol.~45, no.~20, pp. 794--799, 2012.

\bibitem{u1}
U.~Ali, R.~Hafiz, T.~Tauqeer, U.~Younis, W.~Ali, and A.~Ahmad, ``Towards machine learning based real-time fault identification and classification in high power induction motors,'' in \emph{2020 5th International Conference on Robotics and Automation Engineering (ICRAE)}, 2020, pp. 46--53.

\bibitem{u3}
\BIBentryALTinterwordspacing
U.~Ali, ``Towards fault diagnosis in induction motor using fractional fourier transform,'' 2024. [Online]. Available: \url{https://arxiv.org/abs/2412.18227}
\BIBentrySTDinterwordspacing

\bibitem{u4}
U.~Ali, W.~Ali, M.~U. Noor, M.~Umer~Ramzan, M.~U. Aslam, and H.~Farooq, ``Test rig for the fault diagnosis of 3-phase small scale induction motor,'' in \emph{2023 International Conference on IT and Industrial Technologies (ICIT)}, 2023, pp. 1--5.

\bibitem{b10}
Y.~Xie, J.~Guo, P.~Chen, and Z.~Li, ``Coupled fluid-thermal analysis for induction motors with broken bars operating under the rated load,'' \emph{Energies}, vol.~11, no.~8, p. 2024, 2018.

\bibitem{b11}
D.~Zhen, Z.~Wang, H.~Li, H.~Zhang, J.~Yang, and F.~Gu, ``An improved cyclic modulation spectral analysis based on the cwt and its application on broken rotor bar fault diagnosis for induction motors,'' \emph{Applied Sciences}, vol.~9, no.~18, p. 3902, 2019.

\bibitem{b13}
R.~S. Kumar, I.~G.~C. Raj, I.~Alhamrouni, S.~Saravanan, N.~Prabaharan, S.~Ishwarya, M.~Gokdag, and M.~Salem, ``A combined ht and ann based early broken bar fault diagnosis approach for ifoc fed induction motor drive,'' \emph{Alexandria Engineering Journal}, vol.~66, pp. 15--30, 2023.

\bibitem{b14}
M.~E. E.-D. Atta, D.~K. Ibrahim, and M.~I. Gilany, ``Broken bar fault detection and diagnosis techniques for induction motors and drives: State of the art,'' \emph{IEEE Access}, vol.~10, pp. 88\,504--88\,526, 2022.

\bibitem{b16}
V.~Fernandez-Cavero, J.~Pons-Llinares, O.~Duque-Perez, and D.~Morinigo-Sotelo, ``Detection and quantification of bar breakage harmonics evolutions in inverter-fed motors through the dragon transform,'' \emph{ISA transactions}, vol. 109, pp. 352--367, 2021.

\bibitem{b17}
W.~F. Godoy, I.~N. da~Silva, A.~Goedtel, R.~H.~C. Pal{\'a}cios, and T.~D. Lopes, ``Application of intelligent tools to detect and classify broken rotor bars in three-phase induction motors fed by an inverter,'' \emph{IET Electric Power Applications}, vol.~10, no.~5, pp. 430--439, 2016.

\bibitem{b18}
O.~Yaman, ``An automated faults classification method based on binary pattern and neighborhood component analysis using induction motor,'' \emph{Measurement}, vol. 168, p. 108323, 2021.

\bibitem{b19}
A.~Kerboua, A.~Metatla, R.~Kelailia, and M.~Batouche, ``Fault diagnosis in induction motor using pattern recognition and neural networks,'' in \emph{2018 International Conference on Signal, Image, Vision and their Applications (SIVA)}.\hskip 1em plus 0.5em minus 0.4em\relax IEEE, 2018, pp. 1--7.

\bibitem{b20}
X.~Li, Y.~Li, K.~Yan, H.~Shao, and J.~J. Lin, ``Intelligent fault diagnosis of bevel gearboxes using semi-supervised probability support matrix machine and infrared imaging,'' \emph{Reliability Engineering \& System Safety}, vol. 230, p. 108921, 2023.

\bibitem{b21}
S.~Shao, S.~McAleer, R.~Yan, and P.~Baldi, ``Highly accurate machine fault diagnosis using deep transfer learning,'' \emph{IEEE Transactions on Industrial Informatics}, vol.~15, no.~4, pp. 2446--2455, 2018.

\bibitem{b22}
S.~M.~K. Zaman and X.~Liang, ``An effective induction motor fault diagnosis approach using graph-based semi-supervised learning,'' \emph{IEEE Access}, vol.~9, pp. 7471--7482, 2021.

\bibitem{b24}
O.~D. Sanchez, G.~Martinez-Soltero, J.~G. Alvarez, and A.~Y. Alanis, ``Real-time neural classifiers for sensor faults in three phase induction motors,'' \emph{IEEE Access}, vol.~11, pp. 19\,657--19\,668, 2023.

\bibitem{b25}
S.~Misra, S.~Kumar, S.~Sayyad, A.~Bongale, P.~Jadhav, K.~Kotecha, A.~Abraham, and L.~A. Gabralla, ``Fault detection in induction motor using time domain and spectral imaging-based transfer learning approach on vibration data,'' \emph{Sensors}, vol.~22, no.~21, p. 8210, 2022.

\bibitem{b23}
K.~Barrera-Llanga, J.~Burriel-Valencia, {\'A}.~Sapena-Ba{\~n}{\'o}, and J.~Mart{\'\i}nez-Rom{\'a}n, ``A comparative analysis of deep learning convolutional neural network architectures for fault diagnosis of broken rotor bars in induction motors,'' \emph{Sensors}, vol.~23, no.~19, p. 8196, 2023.

\bibitem{b26}
\BIBentryALTinterwordspacing
A.~Elly~Treml, R.~Andrade~Flauzino, M.~Suetake, and N.~A. Ravazzoli~Maciejewski, ``Experimental database for detecting and diagnosing rotor broken bar in a three-phase induction motor.'' 2020. [Online]. Available: \url{https://dx.doi.org/10.21227/fmnm-bn95}
\BIBentrySTDinterwordspacing

\bibitem{b27}
H.~K. Kwok and D.~L. Jones, ``Improved instantaneous frequency estimation using an adaptive short-time fourier transform,'' \emph{IEEE transactions on signal processing}, vol.~48, no.~10, pp. 2964--2972, 2000.

\bibitem{b28}
S.~Sayyad, S.~Kumar, A.~Bongale, K.~Kotecha, G.~Selvachandran, and P.~N. Suganthan, ``Tool wear prediction using long short-term memory variants and hybrid feature selection techniques,'' \emph{The International Journal of Advanced Manufacturing Technology}, vol. 121, no. 9-10, pp. 6611--6633, 2022.

\bibitem{b29}
S.~Do, K.~D. Song, and J.~W. Chung, ``Basics of deep learning: a radiologist's guide to understanding published radiology articles on deep learning,'' \emph{Korean journal of radiology}, vol.~21, no.~1, pp. 33--41, 2020.

\end{thebibliography}

\end{document}